\documentclass[runningheads]{llncs}

\usepackage[final,year=2026,ID=9180]{eccv}
\usepackage{eccvabbrv}

\usepackage[accsupp]{axessibility}  % Improves PDF readability for those with disabilities.
\definecolor{LightCyan}{rgb}{0.88,1,1}
\usepackage[pagebackref,breaklinks,colorlinks,citecolor=eccvblue]{hyperref}
\usepackage{graphicx}
\usepackage{booktabs}
\usepackage{fix-cm}
\usepackage{bm}
\usepackage{multirow}
\usepackage{longtable}
\usepackage{supertabular}
\usepackage{colortbl}
\usepackage{subcaption}
\usepackage[most]{tcolorbox}
\usepackage[dvipsnames]{xcolor}
\usepackage{wrapfig}
\usepackage{adjustbox}
\usepackage{array}
\usepackage{tcolorbox} \tcbuselibrary{breakable}
\usepackage{xspace}

\usepackage{orcidlink}

\newcommand{\myPara}[1]{\vspace{5pt}\noindent$\bullet$~\textbf{#1} \ }
\newcommand{\supp}[1]{\textcolor{magenta}{#1}}

\newcommand\blfootnote[1]{
    \begingroup
    \renewcommand\thefootnote{}\footnote{#1}
    \addtocounter{footnote}{-1}
    \endgroup
}

\begin{document}

% ---------------------------------------------------------------
\def\model{Omni-o3\xspace}
\def\dataset{OmniReason\xspace}
\title{\model: Deep Nested Omnimodal Deduction\\for Deliberative Audio-Visual Reasoning}

\titlerunning{\model}
\authorrunning{Zhang et al.}
\author{
\textbf{Zhicheng Zhang$^{1}$, Wentao Gu$^{1}$, Weicheng Wang$^{1}$, Yongjie Zhu$^{3\dag}$}\\
\textbf{Wenyu Qin$^3$, Meng Wang$^3$, Pengfei Wan$^3$, Jufeng Yang$^{124\ddag}$}\\
} 

\institute{{\small $^1$ Nankai University  $^2$ Pengcheng Laboratory  $^3$ Kuaishou Technology}\\
{\small $^4$ Nankai International Advanced Research Institute (SHENZHEN·FUTIAN)}\\
{\scriptsize
\texttt{gloryzzc6@sina.com,jimgu2004@163.com,wangweicheng777@mail.nankai.edu.cn}}\\
{\scriptsize
\texttt{\{zhuyongjie,qinwenyu,wangmeng46,wanpengfei\}@kuaishou.com,yangjufeng@nankai.edu.cn
}}
}

% Second Author\inst{2,3}\orcidlink{1111-2222-3333-4444} \and
% Third Author\inst{3}\orcidlink{2222--3333-4444-5555}}


% Springer Heidelberg, Tiergartenstr.~17, 69121 Heidelberg, Germany

{
\maketitle
\begin{figure}[h]
\centering
% \vspace{-30pt}
\includegraphics[width=\linewidth]{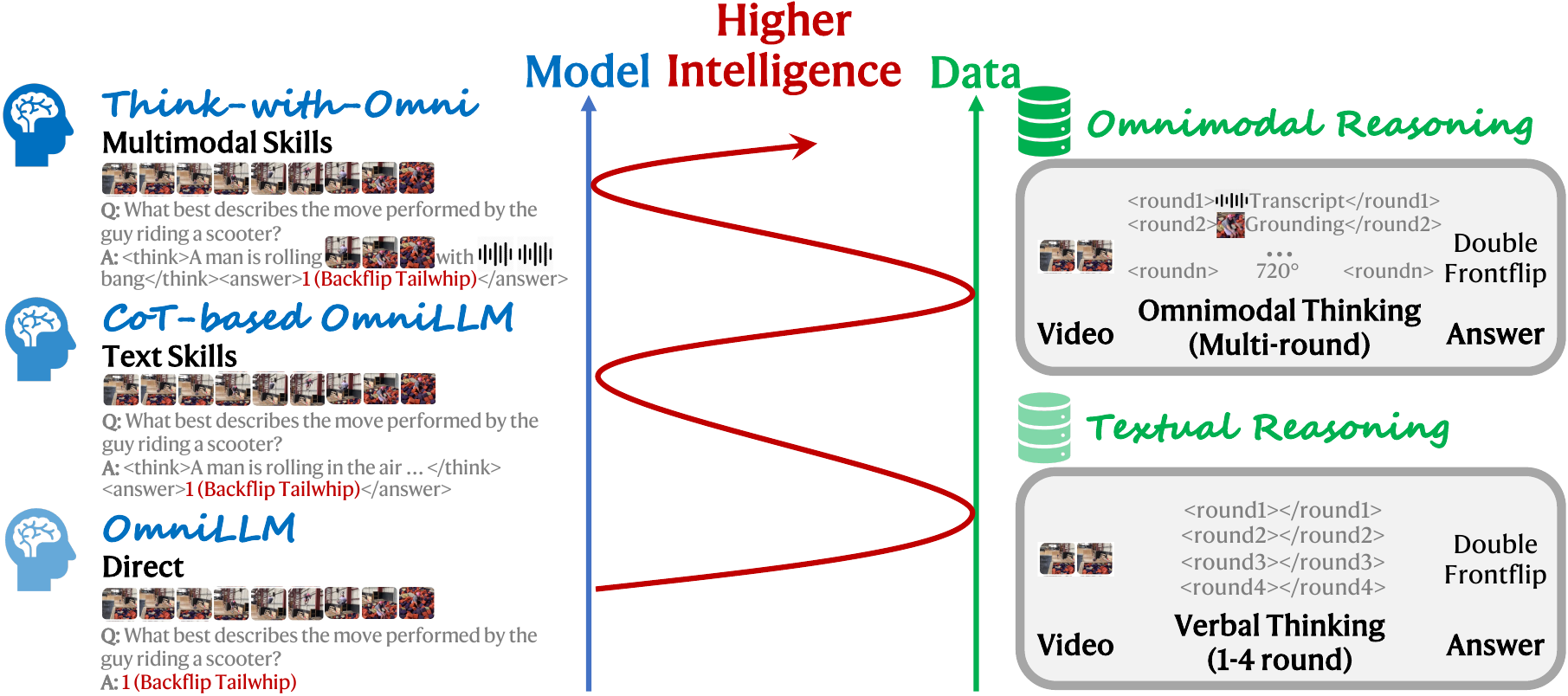}
% \vspace{-20pt}
\caption{
\textbf{Paradigm Comparison.} 
We propose \model driven by \textit{Think-with-Omni}.
By embedding omni skills in multi-round deduction, it overcomes direct-response and verbal CoT, elevating shallow verbal thinking to deliberative omnimodal reasoning.
}
% \vspace{-40pt}
\label{fig:Motivation}
\end{figure}
}
\blfootnote{$\dag$: Project Leader.  $\ddag$: Corresponding Author.}
\begin{abstract}
Omnimodal understanding entails a massive, highly redundant search space of cross-modal interactions, demanding focused and deliberative reasoning.
Current reasoning paradigms rely on either sequential step-by-step generation or parallel sample-by-sample rollouts, leading to isolated reasoning trajectories.
This inability to share promising intermediate paths severely limits exploration efficiency and causes compounding errors in complex audio-visual tasks. 
To break this bottleneck, we introduce \model, a novel framework driven by a \textit{deep nested deduction} policy. 
By formulating reasoning as a dynamic recursive search, \model inherently shares reasoning prefixes across branches, enabling the iterative execution of four atomic cognitive actions: expansion, selection, simulation, and backpropagation. 
To empower this framework, we propose a robust two-stage training paradigm: 
(1) cold-start supervised fine-tuning on 101K high-quality, long-chain trajectories distilled from 3.5M diverse omnimodal samples, enabling necessary recursive search patterns; 
and (2) nested group rollout-driven exploratory reinforcement learning on 18K complex multi-turn samples, explicitly guided by a novel multi-step reward model to stimulate deep nested reasoning. 
Extensive experiments demonstrate that \model achieves competitive performance across 11 benchmarks, unlocking advanced capabilities in comprehensive audio-visual, visual-centric, and audio-centric reasoning tasks.
\textbf{Code, data, and model are available on \supp{HomePage}.}
\end{abstract}
\section{Introduction}

Multimodal Large Language Models (MLLMs), designed to process and reason over interleaved audio, visual, and textual streams, are widely recognized as a crucial stepping stone toward artificial general intelligence~\cite{hendrycks2025definition,spearman1961general} and embodied agents~\cite{fung2025embodied,szot2025multimodal}.
As a natural extension of text-centric foundation models, recent research has embarked on preliminary explorations in the omnimodal domain, evolving from basic instruction following to complex reasoning~\cite{o3,openai2025o3-mini}.
Leveraging the massive textual knowledge embedded during LLM pre-training, these advanced models increasingly utilize reinforcement learning (RL) to elicit emergent reasoning behaviors~\cite{shao2024deepseekmath}.
For instance, AVATAR~\cite{kulkarni2025avatar} proposes an off-policy RL framework with Temporal Advantage Shaping, while Omni-R1~\cite{zhong2025omni} employs a two-system architecture with hierarchical rewards.
All of these methods relies on reasoning paradigms, which can generally be categorized into two distinct approaches: sequential and parallel.
Sequential reasoning~\cite{wang2022self,wei2022chain} relies on step-by-step generation, which inherently deepens the search process and allows for rigorous, multi-hop logical deduction.
Conversely, parallel reasoning~\cite{fu2025deep,ning2023skeleton} involves sample-by-sample rollouts, significantly improving exploration diversity and preventing the model from being trapped in local optima.
Together, these paradigms make the progressive scale-up of model intelligence possible, echoing the monumental success observed in pure-text reasoning models (\eg, OpenAI o3, Gemini 3.1 Pro).

However, as the scale of reasoning increases, the search space and computational overhead of omnimodal LLMs escalate exponentially~\cite{hong2025deepeyesv2,zheng2025deepeyes}.
Unlike pure text tasks, a seemingly simple omnimodal QA instruction can easily span up to 128K tokens due to the dense nature of continuous signals~\cite{gao2025apvr,shu2025video}.
More critically, the ratio of omnimodal tokens (\ie, visual patches and audio spectrograms) to textual tokens often exceeds 100:1 such as Video-MME~\cite{fu2025video}.
This massive, highly redundant cross-modal context creates an insurmountable chasm for directly replicating the successful scaling laws and reasoning paradigms of text-based LLMs.
Under existing sequential or parallel paradigms, the inability to share intermediate visual-audio processing paths leads to severe computational inefficiency, isolated reasoning trajectories, and compounding errors.

To bridge this chasm, we introduce \textbf{\model}, a novel framework driven by a \textit{deep nested deduction} policy.
Instead of relying on isolated step-by-step generation or parallel rollouts—which force models to independently and redundantly encode the massive audio-visual context for every single trajectory—\model formulates reasoning as a dynamic recursive search.
Specifically, our nested deduction structures the reasoning process into a hierarchical tree.
By inherently sharing the heavy omnimodal reasoning prefixes (the ``roots'') across multiple exploratory branches (the ``leaves''), \model only computes the dense visual and audio tokens once.
This mechanism drastically reduces the redundancy and shifts the computational burden away from repetitive perception encoding, focusing it entirely on deliberative cognitive exploration to effectively bypass the 100:1 token imbalance.
Built upon this shared-prefix architecture, our framework enables the iterative execution of four atomic cognitive actions: \textit{selection, expansion, simulation, and backpropagation}.

To empower this framework, we propose a robust two-stage training paradigm specifically tailored for omnimodal deduction:
First, we conduct a cold-start supervised fine-tuning (SFT) on 101K distilled long-chain trajectories.
Rather than simply imitating final answers, this stage enables the model to internalize the structural syntax of nested reasoning (\ie, hierarchical formatting via \texttt{\# Turn} and \texttt{\#\# Action} tags) and bootstrap its intrinsic omnimodal capabilities.
By doing so, it transforms the model's implicit multimodal knowledge into explicit, executable cognitive skills (\eg, internal audio transcription, spatial grounding, and temporal localization), laying the foundation for multi-turn self-verification.
Second, to overcome the exposure bias inherent in SFT and unlock true autonomous exploration, we introduce a nested group rollout-driven exploratory reinforcement learning phase on 18K complex multi-turn samples.
In this stage, the model dynamically explores diverse reasoning trajectories formulated as tree search.
To effectively supervise this massive search space, we design a novel multi-step reward modeling mechanism that provides fine-grained feedback across intermediate reasoning prefixes, child node preferences, structural formatting, and final outcomes.
This explicitly guides the model to dynamically assess intermediate states, backtrack from dead ends, and iteratively refine its deductions, ultimately stimulating deep and verifiable reasoning.

To support this rigorous training pipeline and address the scarcity of deliberative audio-visual data, we develop an automated data curation engine driven by Monte Carlo Tree Search (MCTS).
Starting from a vast pool of 3.5M omnimodal samples, the engine iteratively constructs reasoning trees by simulating and evaluating intermediate cognitive actions.
Through this process, it produces 101K high-quality, long-horizon, tree-structured reasoning trajectories for SFT,
and further curates 18K complex, diverse, and challenging samples for RL according to the number of turns required to solve each problem.
Unlike traditional single-turn question-answering datasets, our curated dataset encompasses diverse modality combinations (\eg, Audio + Visual) and features deep conversational structures spanning up to 10+ turns.
This structural depth provides the essential multi-turn interactions and dense reward signals required to bridge the gap between raw perception and system-2 multimodal reasoning~\cite{xiang2025towards}.

The main contributions of this work are two-fold and as follows:
(1) We propose \model, an omnimodal deduction framework that formulates complex reasoning as a dynamic tree search.
Experimental results show that \model achieves state-of-the-art performance compared to existing models across diverse video reasoning, quality assessment, and temporal localization benchmarks.
(2) We present a high-quality audio-visual reasoning dataset comprising 101K SFT multi-turn trajectories and 18K difficult \& diverse RL samples.
We believe it can serve as a comprehensive data infrastructure for advancing traceable and grounded omnimodal deduction.

\section{Related Work}
\noindent\textbf{Multimodal Large Language Model (MLLM).}
Multimodal Large Language Models (MLLMs) have garnered significant attention recently due to their ability to integrate pre-trained foundational models, especially powerful Large Language Models (LLMs)~\cite{achiam2023gpt,touvron2023llama}, alongside multimodal encoders~\cite{dosovitskiy2021an,radford2021learning}.
The paradigm of multimodal learning has transitioned from modular, perception-driven pipelines~\cite{alayrac2022flamingo, li2023blip} to unified foundation models that scale both vision and language components~\cite{bai2025qwen3,chen2024expanding}, significantly enhancing the processing of multimodal inputs and outputs~\cite{alayrac2022flamingo,bai2023qwen}.
MMVP~\cite{tong2024eyes} identifies that existing MLLMs often fail to fully activate the vision modality due to improper handling of visual attributes.
Furthermore, Cambrian-1~\cite{tong2024cambrian} confirms this limitation and introduces a spatial vision aggregator to enhance visual features.
This indicates a pressing need for paradigms like our Omni-o3, which goes beyond mere perception scaling to deeply internalize and verify omnimodal cues during the deliberative reasoning process.
%
% Recent state-of-the-art models, such as InternVL 2.5 and 3.5~\cite{chen2024expanding, wang2025internvl3}, utilize a ViT-MLP-LLM architecture with progressive scaling and a visual resolution router to match the performance of proprietary systems.
% %
% Similarly, Qwen3-VL~\cite{bai2025qwen3} introduces interleaved-MRoPE for simultaneous spatial-temporal modeling, supporting massive contexts for long-video understanding, while DeepSeek-VL2~\cite{wu2024deepseek} integrates sparse mixture-of-experts with multi-head latent attention for highly efficient long-sequence inference. 
% %
% Despite these architectural advancements, the intrinsic potential of multimodal inputs remains underutilized in many systems.

\noindent\textbf{Reasoning Model in MLLM.}
With the blossom of a series of recent models such as Gemini Pro-series and OpenAI o-series~\cite{comanici2025gemini,guo2025deepseek,openai2025o3-mini}, various works probe into integrating MLLMs with reasoning capacity~\cite{ahn-etal-2024-large}.
Multimodal chain-of-thought (MCoT) prompting~\cite{li2025imagine,liu2024ocrbench,thawakar2025llamav} offers a step-by-step reasoning trajectory when MLLM faces hard questions including detail grounding~\cite{liu2024mminstruct,wu2024v}, agent planing~\cite{li2025imagine}, \etc.
Specifically, MCoT decomposes the question into a group of reasoning steps and builds a chain to guide the model to generate the results of complex problems step-by-step~\cite{qwen2024qvq,xiang2025towards,zhang2024improve}.
Recent works have demonstrated that CoT prompting substantially improves the MLLM’s capability on reasoning tasks.
For instance, LLaVA-CoT~\cite{xu2024llava} prompts MLLM reasoning steps into the summary, caption, reasoning, and conclusion stages and proposes a stage-level beam search strategy to further enhance reasoning capacity.
LLaVA-Reasoner~\cite{xu2025llava} pioneers the use of forced Chain-of-Thoughts, establishing a new direction for structured prompting techniques.
In this paper, we propose spatial-temporal cues enhanced reasoning as intermediate bridge to achieve confidential video understanding.

\noindent\textbf{Omni-Modal MLLM.}
Recent advanced works in audio-visual understanding have been significantly bolstered by the development of innovative datasets and models.
AURELIA~\cite{chowdhury2025aurelia} introduces a novel actor-critic reasoning framework that distills structured, step-by-step reasoning into audio-visual LLM at test time. 
To evaluate such capabilities in daily scenarios, Daily-Omni~\cite{zhou2025daily} provides a dedicated benchmark alongside a training-free agent baseline. 
Recent research has increasingly leveraged RL to elicit emergent reasoning behaviors; for instance, EchoInk-R1~\cite{xing2025echoink} utilizes group relative policy optimization (GRPO) for cross-modal alignment, while 
HumanOmniV2~\cite{yang2025humanomniv2} introduces IntentBench and logical rewards to address shortcut problems through explicit context summarization. 
Furthermore, Omni-R1~\cite{zhong2025omni} employs a two-system architecture with hierarchical rewards to balance temporal coverage and spatial resolution, and AVATAR~\cite{kulkarni2025avatar} proposes an off-policy RL framework with Temporal Advantage Shaping (TAS) for improved credit assignment. 
To address the scarcity of multimodal reasoning data, AVRT~\cite{araujo2025avrt} introduces a novel reasoning transfer framework that distills and merges reasoning traces from specialized single-modality (audio and visual) teachers via a text-only LLM, generating high-quality cross-modal reasoning data to facilitate effective SFT and RL.

\section{Methodology}

\begin{wrapfigure}{r}{.55\textwidth}
    \centering
    \vspace{-28pt} % 如果图片上方空白太多，可以用负值调整
    \includegraphics[width=\linewidth]{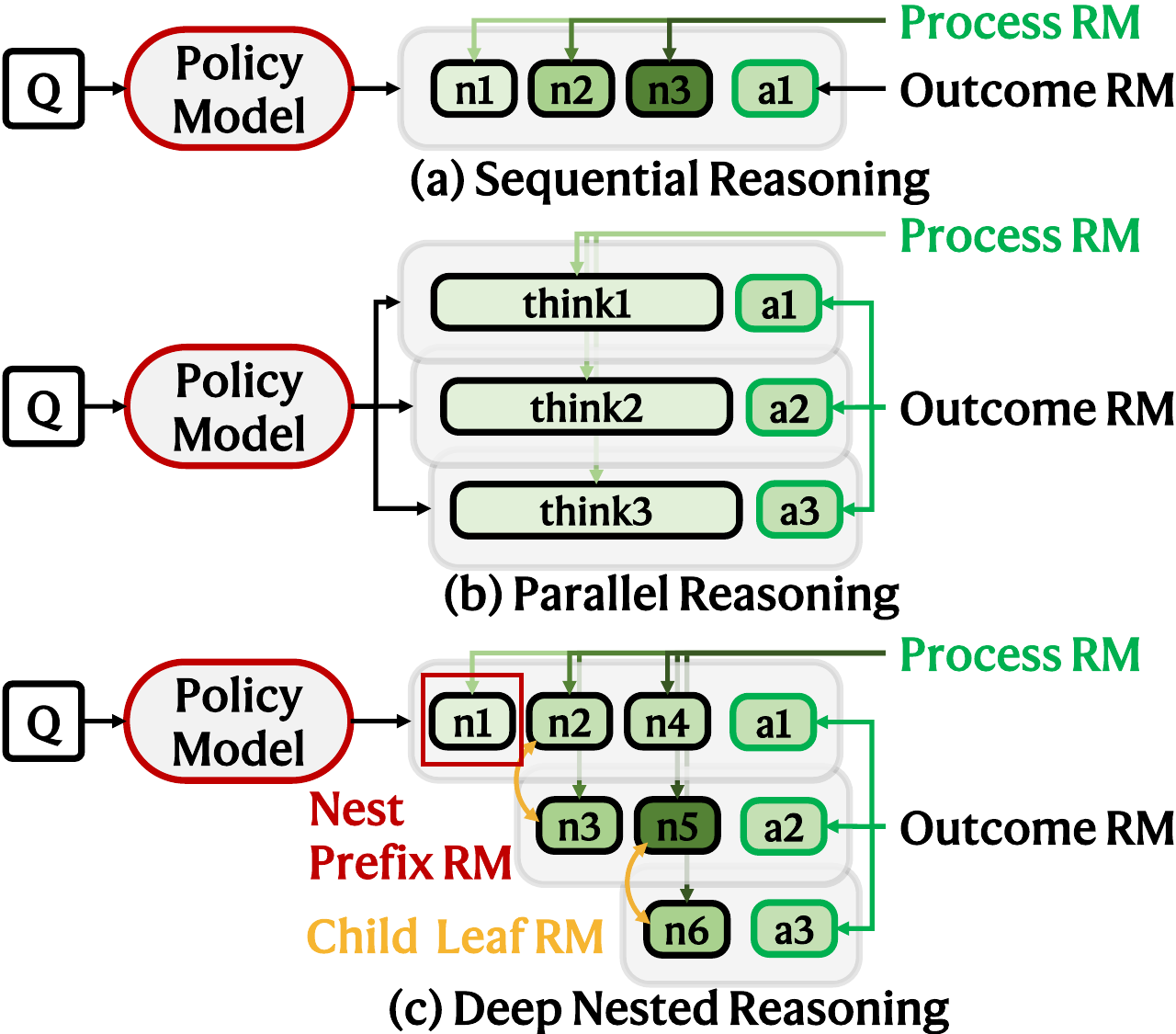}
    \vspace{-18pt}
    \caption{
    \textbf{Reasoning paradigms.}
From top to bottom: CoT, BoN, and our nested reasoning.
%     \textbf{Illustration of different reasoning paradigms.}
% From top to bottom: standard CoT reasoning (linear generation), GRPO-style reasoning (parallel independent sampling), and our proposed nested reasoning (iterative refinement).
% Unlike previous methods, nested reasoning leverages tool-use feedback loops (blue arrows) and hierarchical answer aggregation (green arrows) to handle complex multimodal tasks.
    }
    \label{fig:reasoning_paradigms}
    \vspace{-15pt}
\end{wrapfigure}
\subsection{Preliminary}
Let $\mathcal{V}$ denote the visual input (\eg, a video clip or a set of images) and $\mathcal{Q}$ denote the textual query.
The goal of a multimodal reasoning model is to generate an optimal answer $y^*$ that maximizes the probability $P(y|\mathcal{V}, \mathcal{Q})$.
We denote the model (\eg, a Large Vision-Language Model) as $\pi_\theta$, parameterized by $\theta$. 
We categorize existing reasoning approaches into three paradigms, as illustrated in Figure~\ref{fig:reasoning_paradigms}.

\myPara{Sequential Reasoning.} 
Standard CoT approaches decompose the reasoning process into a linear sequence of intermediate thought steps $z = \{z_1, z_2, \dots, z_T\}$.
The model generates the final answer $y$ conditioned on the input and the chain of thoughts:
\begin{equation}
    P_{CoT}(y|\mathcal{V}, \mathcal{Q}) = \prod_{t=1}^T \pi_\theta(z_t | \mathcal{V}, \mathcal{Q}, z_{<t}) \cdot \pi_\theta(y | \mathcal{V}, \mathcal{Q}, z)
\end{equation}
While effective for textual tasks, standard CoT (Figure~\ref{fig:reasoning_paradigms}, top) often struggles with complex visual grounding, as it lacks mechanisms to interact with the visual content dynamically.

\myPara{Parallel Reasoning.}
To enhance robustness, recent methods (\eg, Self-Consistency, GRPO) sample multiple independent reasoning paths in parallel.
Let $\{z^{(k)}\}_{k=1}^K$ be a set of $K$ independent reasoning trajectories sampled from $\pi_\theta$. 
Each trajectory produces a candidate answer $y^{(k)}$:
\begin{equation}
    y^{(k)} \sim \pi_\theta(\cdot | \mathcal{V}, \mathcal{Q}, z^{(k)}), \quad k \in \{1, \dots, K\}
\end{equation}
The final answer is typically derived via majority voting or a reward-based selection mechanism (Figure~\ref{fig:reasoning_paradigms}, middle). 
However, these paths are isolated; failures in one branch cannot inform or correct other branches.

% In this section, we present an overview of \model, a unified framework designed to address complex reasoning tasks: {Audio-Visual Reasoning}, {Video Reasoning}, and {Audio Reasoning}.
% %
% As illustrated in Figure~\ref{fig:overview}, our methodology deviates from traditional label-fitting approaches by adopting a {Data-Model Co-Evolution Paradigm}, which integrates a comprehensive data construction pipeline with a System-2 reasoning architecture.

\subsection{Deep Nested Deduction: A Tree-Search Formulation}
\label{sec:nested_deduction}
To overcome the limitations of linear and isolated reasoning paradigms, we formulate the deliberative reasoning process as a dynamic tree search, termed {Deep Nested Deduction} (Figure~\ref{fig:reasoning_paradigms}c). 
Instead of generating a flat sequence of tokens, our model $\pi_\theta$ constructs a reasoning tree $\mathcal{T}$. At each step, the model generates a node $n_t$, which encapsulates an atomic cognitive action: \textit{expansion} (proposing new thoughts or sub-goals), \textit{selection} (choosing the most promising branch), \textit{simulation} ({executing the model's intrinsic omnimodal skills to gather internal observations}), or \textit{backpropagation} (updating node values).
Let $\mathcal{P}(n_t)$ denote the trajectory path from the root to node $n_t$. The generation of a nested trajectory is formulated as:
\begin{equation}
    P_{Nest}(\mathcal{T}|\mathcal{V}, \mathcal{Q}) = \prod_{n_t \in \mathcal{T}} \pi_\theta \big(n_t \big| \mathcal{V}, \mathcal{Q}, \mathcal{P}(n_t) \big)
\end{equation}
Unlike standard CoT, this nested formulation allows the model to dynamically assess intermediate outcomes, backtrack from dead ends, and iteratively refine its deductions through multi-turn self-interactions.

\begin{figure*}[!tbp]
  \centering
  % \vspace{20pt}
  \includegraphics[width=.95\linewidth]{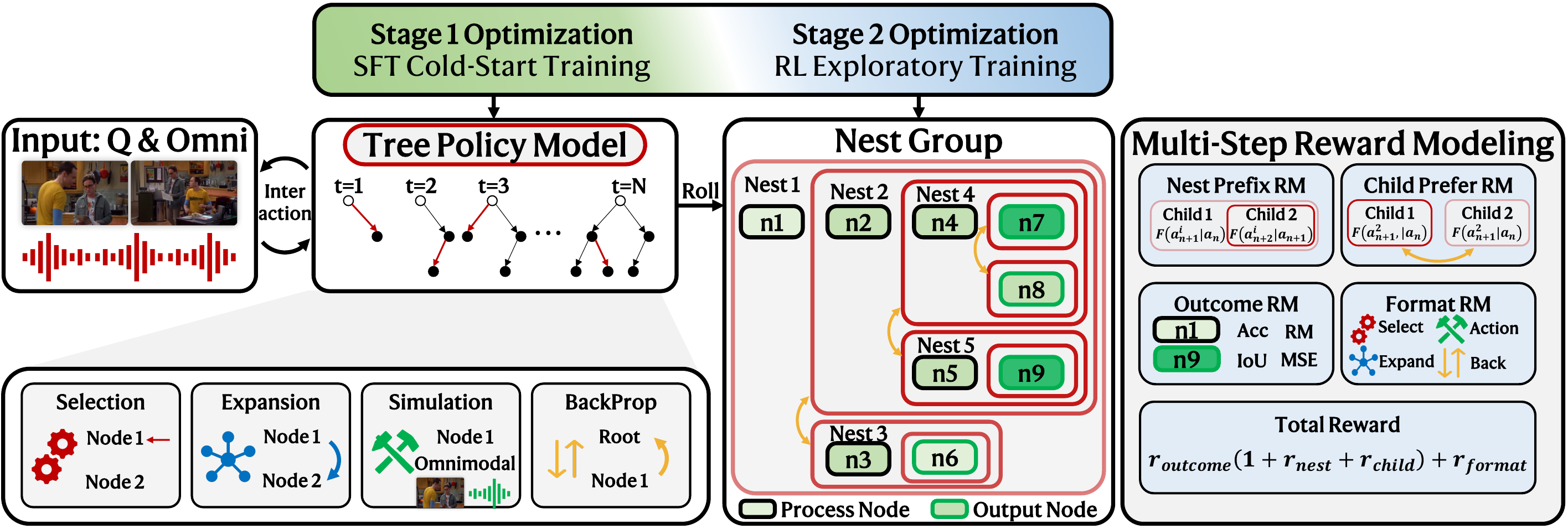}
  \caption{
\textbf{Overall pipeline of the proposed Deep Nested Deduction framework}.
The training paradigm is decoupled into two progressive optimization stages.
Stage 1 SFT Cold-Start Training: The policy model is fine-tuned to internalize the structural syntax of nested reasoning and bootstrap its intrinsic omnimodal capabilities.
Stage 2 RL Exploratory Training: The model autonomously explores reasoning trajectories formulated as a dynamic tree search
  }
  \label{fig:pipeline}
\end{figure*}
\myPara{Stage I: Bootstrapping Intrinsic Omni-Skills via SFT}
Before engaging in autonomous exploration, the model must internalize the structural syntax of nested reasoning and learn to {self-bootstrap its intrinsic omnimodal capabilities} (\eg, internal audio transcription, spatial grounding, and temporal localization). 
In this stage, we utilize our curated high-quality dataset containing 101K long-chain trajectories (detailed in Section~\ref{sec:data_engine}) to perform SFT. 
Given a ground-truth nested trajectory $\mathcal{T}^* = \{n_1^*, n_2^*, \dots, n_N^*\}$, we optimize the model using the standard auto-regressive cross-entropy loss:
\begin{equation}
    \mathcal{L}_{SFT} = - \sum_{t=1}^{N} \log \pi_\theta \big(n_t^* \big| \mathcal{V}, \mathcal{Q}, n_{<t}^* \big)
\end{equation}
This cold-start training transforms the model's implicit multimodal knowledge into explicit, executable cognitive skills, equipping it with the foundational capability to format its outputs into hierarchical structures (\eg, \texttt{<\# Turn>}, \texttt{<\#\# Action>}) and invoke self-verification seamlessly.

\myPara{Stage II: Deliberative Exploration via Multi-Step RL}
While SFT imparts structural knowledge, it suffers from exposure bias and limits the model's ability to discover novel reasoning paths. To stimulate deeper intelligence, we introduce an exploratory reinforcement learning stage using 18K difficult and diverse samples.
As illustrated in Figure~\ref{fig:pipeline} (Right), we design a comprehensive multi-step reward modeling mechanism to provide dense signals for the nested structures. The total reward $r_{total}$ is defined as a synergistic combination of outcome, process, and format rewards:
\begin{equation}
    r_{total} = r_{outcome} \cdot (1 + r_{nest} + r_{child}) + r_{format}
\end{equation}
Specifically:
\begin{itemize}
    \item \textbf{Outcome RM ($r_{outcome}$):} Evaluates the final answer's correctness using metrics such as Accuracy, IoU, or MSE, depending on the task type.
    \item \textbf{Process RM ($r_{nest}, r_{child}$):} Provides fine-grained supervision for intermediate steps. $r_{nest}$ assesses the quality of the nest prefix, while $r_{child}$ evaluates the logic of child leaf nodes, encouraging effective exploration and penalizing logical hallucinations.
    \item \textbf{Format RM ($r_{format}$):} A rule-based penalty that ensures the model strictly adheres to the predefined tree-search syntax (\eg, correct tags for \textit{Select}, \textit{Action}, \textit{Expand}, \textit{Back}).
\end{itemize}

\begin{figure*}[!tbp]
  \centering
    % \vspace{20pt}

  \includegraphics[width=.95\linewidth]{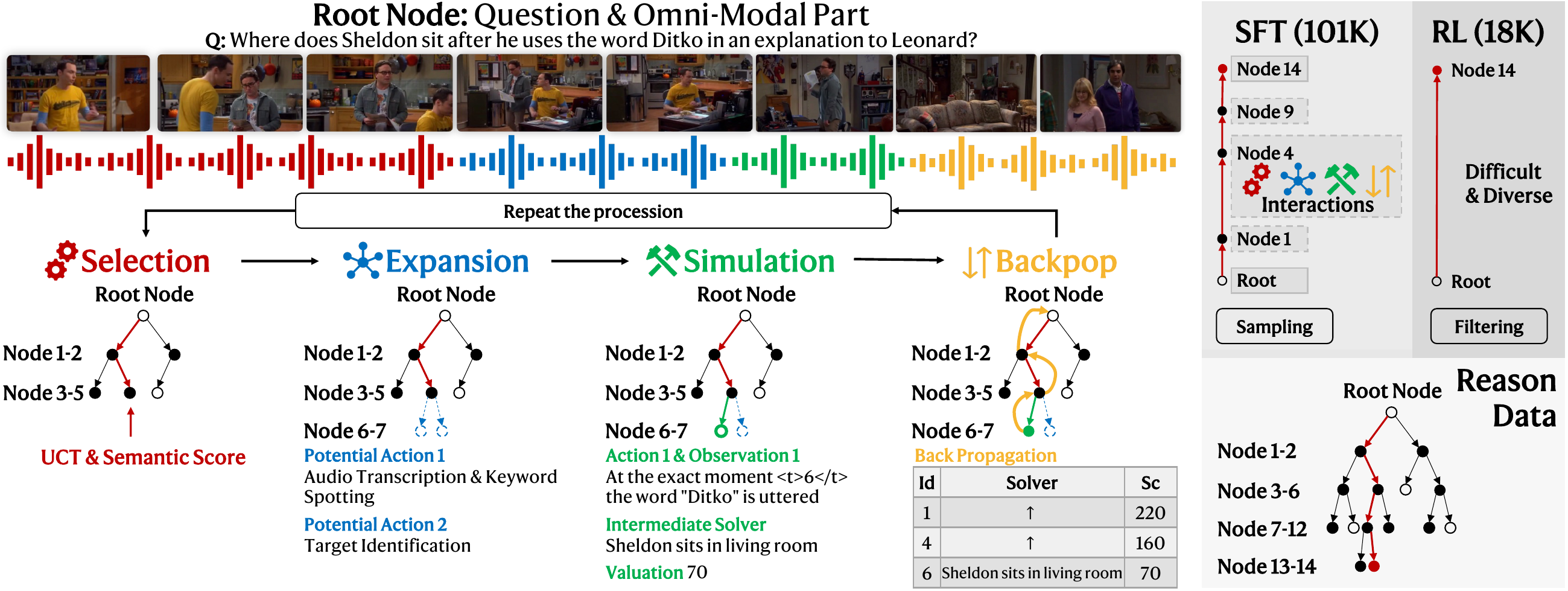}
  \caption{
\textbf{Automated Data Engine for Curating Deliberative Trajectories.}
To overcome the scarcity of multi-turn audio-visual reasoning data, we propose a MCTS-driven curation pipeline.
Given an omnimodal root node, the engine iteratively constructs a reasoning tree via four core operations: selection, expansion, simulation, and backpropagation.
Ultimately, this process yields a high-quality reasoning dataset, comprising 101K sampled trajectories for SFT and 18K difficult, diverse trajectories filtered for RL.
}
  \label{fig:data_pipeline}
\end{figure*}
\subsection{Data Engine: Curating Deliberative Trajectories}
\label{sec:data_engine}
\myPara{Multi-Turn Reasoning Data Construction}
The efficacy of our two-stage training heavily relies on the quality of the reasoning data. To address the scarcity of deliberative audio-visual data, we design an automated data curation pipeline driven by Monte Carlo Tree Search (MCTS), as depicted in Figure~\ref{fig:data_pipeline}.
Starting from a vast pool of 3.5M omnimodal samples, we iteratively construct reasoning trees for each query. Given an omnimodal root node (\eg, a video coupled with a complex question), the engine expands the tree through four core operations:

(1) \textbf{Selection:} The algorithm traverses the tree from the root to a leaf node, guided by a combination of Upper Confidence Bounds for Trees (UCT) and semantic scores to balance exploration and exploitation.
(2) \textbf{Expansion:} At the selected node, the model proposes multiple potential cognitive actions, such as audio transcription, keyword spotting, or spatial target identification.
(3) \textbf{Simulation:} The proposed actions are executed by leveraging a suite of specialized intermediate solvers. These solvers return concrete observations (\eg, timestamped subtitles or bounding boxes) and compute a valuation score reflecting the action's helpfulness toward the final answer.
(4) \textbf{Backpropagation:} The valuation scores are propagated back up the tree to update the statistics of the parent nodes, refining the search direction for subsequent iterations.

After the MCTS process terminates, we extract two distinct subsets from the generated reasoning trees.
For the {SFT stage}, we sample 101K high-quality, successful reasoning trajectories.
These linear paths encapsulate rich multi-turn interactions and serve as demonstrations for the model to internalize the nested reasoning syntax.
For the {RL stage}, we apply strict filtering criteria to curate 18K highly complex, multi-branch trees.
These difficult and diverse samples are specifically chosen to provide dense reward signals and encourage the model to explore alternative reasoning paths.
Ultimately, this data engine serves as the cornerstone for bridging the gap between raw perception and system-2 reasoning in multimodal models.

\begin{figure*}[!tbp]
  \centering
    % \vspace{20pt}

  \includegraphics[width=\linewidth]{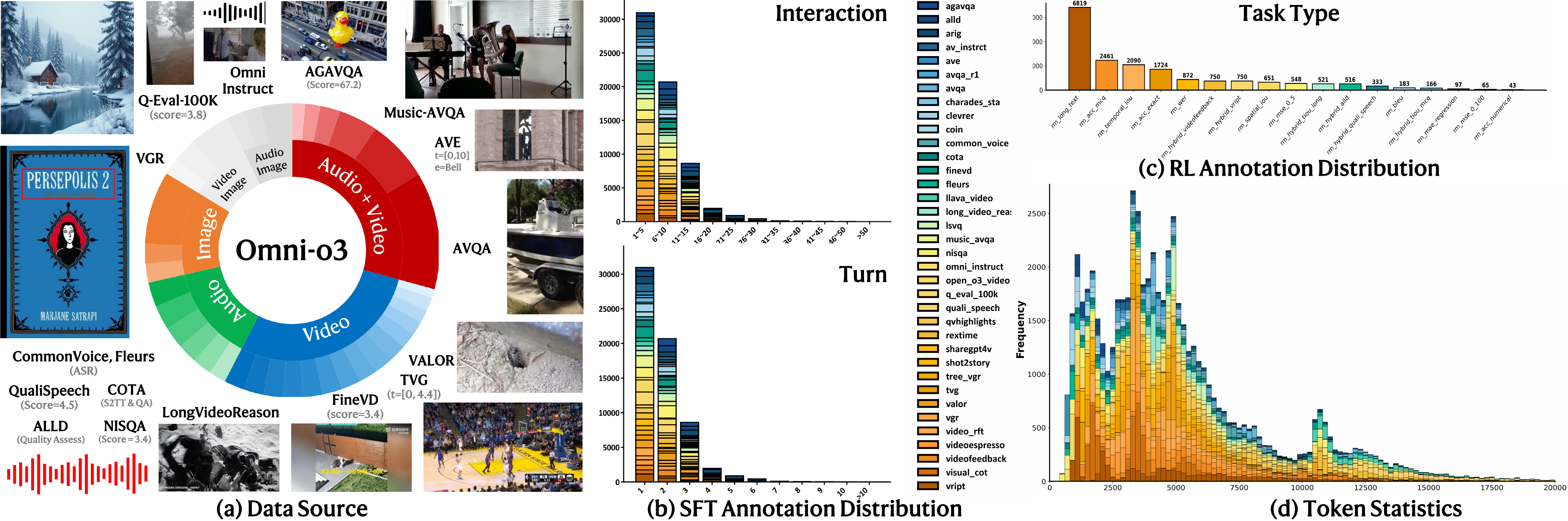}
  \caption{\textbf{Comprehensive statistics of the Omni-o3 training data.}
  We present a detailed overview of our dataset from four key perspectives: {(a) Data Source} illustrates the rich omnimodal composition (Audio, Video, Image, and their combinations) alongside representative sample visualizations. {(b) SFT Annotation Distribution} highlights the complexity of our instruction-tuning data, detailing the frequency of multi-turn dialogues and user-assistant interactions across various sub-datasets. {(c) RL Annotation Distribution} showcases the diverse task types utilized for reward model training, ensuring robust multimodal alignment. Finally, {(d) Token Statistics} demonstrates the long-context nature of our dataset, with a stacked histogram revealing the sequence length distribution across all data sources.}
  \label{fig:datastats}
\end{figure*}
\myPara{Data Statistics}
To provide a comprehensive understanding of the curated Omni-o3 dataset, we present a detailed statistical overview in Figure~\ref{fig:datastats}.
Our analysis highlights the dataset's diversity, complexity, and suitability for training System-2 multimodal models across four key dimensions.

(1) Our dataset is built upon a rich tapestry of multimodal sources, encompassing Audio, Video, Image, and their complex combinations (\eg, Audio+Video, Video+Image).
By integrating diverse sub-datasets such as Music-AVQA, LongVideoReason, and Omni Instruct, we ensure that the model is exposed to a wide spectrum of perceptual challenges, preventing modality collapse and fostering robust omnimodal grounding.
(2) Unlike traditional QA datasets that predominantly feature single-turn, short-form QA pairs, our SFT data is characterized by deep conversational structures.
The distributions of {Interaction} counts and {Turn} lengths demonstrate a heavy-tailed pattern, with a significant proportion of trajectories spanning multiple turns (up to 10+ turns) and complex user-assistant interactions.
This structural depth is crucial for teaching the model the syntax of nested deduction and multi-step self-correction.

(3) For the exploratory RL stage, we curated 18K highly complex reasoning trees across a diverse set of task types.
The distribution illustrates a balanced mix of evaluation metrics, ranging from generative tasks (\eg, \texttt{rm\_long\_text}, \texttt{rm\_bleu}) to precise localization and grounding tasks (\eg, \texttt{rm\_temporal\_iou}, \texttt{rm\_spatial\_iou}, \texttt{rm\_acc\_exact}).
This heterogeneous reward landscape ensures that the model's deliberative exploration is aligned with fine-grained multimodal accuracy rather than mere linguistic fluency.
(4) A hallmark of System-2 reasoning is the ability to maintain coherent thought processes over extended contexts.
The stacked histogram of token statistics reveals that our trajectories consistently demand long-context processing, with sequence lengths frequently exceeding 5,000 tokens and extending up to 20,000 tokens.
This long-context nature forces the model to engage in sustained deliberative thinking, effectively bridging the gap between raw perception and deep cognitive reasoning.

\section{Experiment}
\subsection{Benchmark Datasets}
\begin{table*}[!tbp]
\centering
\tabcolsep=2.8pt
\caption{\textbf{Comparison with leading MLLMs across 8 popular audio-visual benchmarks.}
Bold numbers indicate the best performance.
``All'' denotes the overall score on the respective benchmark.
``N/A'' indicates that the model is either closed-source or doesn't support this task.
}

\begin{minipage}[c]{\linewidth}
\centering
\begin{minipage}[c]{.605\linewidth}
\subfloat[\textbf{General \& Complex QA}]
{%
\centering
\begin{adjustbox}{width=\textwidth}
\begin{tabular}{
p{2.9cm}
p{1.275cm}<{\centering} % VideoMME
p{1.275cm}<{\centering} % WorldSense
p{1.3cm}<{\centering} % Daily-Omni
p{1.45cm}<{\centering} % Video-Holmes
p{1.3cm}<{\centering} % Intent-Bench
}
\toprule
\multirow{2}{*}{Model} & \multicolumn{1}{c}{\textbf{VMME.}} & \multicolumn{1}{c}{\textbf{WSense.}} & \multicolumn{1}{c}{\textbf{DOmni.}} & \multicolumn{1}{c}{\textbf{VHolmes.}} & \multicolumn{1}{c}{\textbf{IntentB.}}\\
\cmidrule(lr){2-2}
\cmidrule(lr){3-3}
\cmidrule(lr){4-4}
\cmidrule(lr){5-5}
\cmidrule(lr){6-6}
& All & All & All & All & All \\
\midrule

\rowcolor{gray!10}
\multicolumn{6}{l}{\textit{Closed-Source APIs}}\\
GPT-4o$_{\color{gray}{\it May\ 24}}$ 
&77.2&42.6&56.5&42.0&60.0\\
GPT-5$_{\color{gray}{\it Dec\ 25}}$ 
&86.0&-   &-   &-   &-   \\
Gemini1.5 Pro$_{\color{gray}{\it Dec\ 24}}$  
&75.0&48.0&-   &45.0&67.2\\
Gemini2.5 Pro$_{\color{gray}{\it May\ 25}}$  
&86.9&-   &-   &-   &-   \\
\midrule

\rowcolor{gray!10}
\multicolumn{6}{l}{\textit{Video MLLMs}}\\
Qwen3-VL$_{\color{gray}{\it Oct\ 25}}$  
&72.5&42.7&57.1&36.1&61.0\\
Qwen2.5-VL$_{\color{gray}{\it Feb\ 25}}$
&65.1&38.3&40.7&27.8&59.7\\
OneThinker$_{\color{gray}{\it Dec\ 25}}$ 
&66.5&42.6&56.9&48.7&59.9\\
Open-o3-video$_{\color{gray}{\it Oct\ 25}}$ 
&63.6&37.5&49.9&35.0&57.7\\
VideoChat-R1.5$_{\color{gray}{\it Sep\ 25}}$ 
&65.2&40.8&53.0&41.0&59.8\\
VideoRFT$_{\color{gray}{\it May\ 25}}$ 
&65.2&39.3&48.5&37.0&58.8\\
VideoChat-R1$_{\color{gray}{\it Apr\ 25}}$ 
&59.8&38.2&48.3&34.0&60.0\\
Video-R1$_{\color{gray}{\it Mar\ 25}}$ 
&59.3&35.5&46.5&36.5&58.5\\
\midrule

\rowcolor{gray!10}
\multicolumn{6}{l}{\textit{Omnimodal MLLMs}}\\
Qwen2.5-Omni$_{\color{gray}{\it Mar\ 25}}$  
&59.8&45.4&47.5&16.4&64.2\\
Qwen3-Omni$_{\color{gray}{\it Sep\ 25}}$  
&70.5&48.3&\textbf{75.8}&50.0&65.1\\
Ola$_{\color{gray}{\it Feb\ 25}}$  
&68.4&44.2&50.7&18.0&57.4\\
EchoInk-R1$_{\color{gray}{\it May\ 25}}$  
&60.8&45.7&46.2&35.0&63.6\\
Omni-R1$_{\color{gray}{\it May\ 25}}$  
&60.7&44.1&46.8&30.0&63.5\\
AV-Reasoner$_{\color{gray}{\it June\ 25}}$  
&56.8&44.6&53.8&39.6&59.6\\
HumanOmniV2$_{\color{gray}{\it June\ 25}}$  
&66.0&47.1&58.5&39.0&\textbf{69.3}\\
AVATAR$_{\color{gray}{\it CVPR\ 26}}$  
&62.8&46.0&55.7&45.1&61.9\\
\midrule

\rowcolor{LightCyan}
\textbf{Omni-o3}  
&\textbf{72.6}&\textbf{53.0}&72.0&\textbf{53.0}&68.4\\
\bottomrule
\end{tabular}%
\end{adjustbox}%
}
\end{minipage}%
% \hspace{1pt}
\hfill
\begin{minipage}[c]{.39\linewidth
}
\subfloat[\textbf{Quality Assessment}]
{%
\begin{minipage}[t]{\linewidth}
\centering
\begin{adjustbox}{width=\textwidth}
\begin{tabular}{
p{3.6cm}
p{2.8cm}<{\centering} % AGAVQA
}
\toprule
\multirow{2}{*}{Model} & \multicolumn{1}{c}{\textbf{AGAVQA}} \\
\cmidrule(lr){2-2}
& SRCC \\
\midrule

\rowcolor{gray!10}
\multicolumn{2}{l}{\textit{Task Specific MLLMs}}\\
% DNN-RNT$_{\color{gray}{\it Mar\ 23}}$
% &.494\\
% DNN-SND$_{\color{gray}{\it Mar\ 23}}$
% &.546\\
% GeneralAVQA$_{\color{gray}{\it Jul\ 23}}$
% &.601\\
AGAV-Rater$_{\color{gray}{\it Jan\ 25}}$
&.746\\
\midrule

\rowcolor{gray!10}
\multicolumn{2}{l}{\textit{Omnimodal MLLMs}}\\
Qwen2.5-Omni$_{\color{gray}{\it Mar\ 25}}$
&.224\\
Qwen3-Omni$_{\color{gray}{\it Sep\ 25}}$
&.486\\
NextGPT$_{\color{gray}{\it Sep\ 23}}$
&.042\\
PandaGPT$_{\color{gray}{\it May\ 23}}$
&.127\\
VITA-1.0$_{\color{gray}{\it Aug\ 24}}$ 
&.128\\
VideoLLaMA2$_{\color{gray}{\it Jul\ 24}}$ 
&.244\\
\midrule

\rowcolor{LightCyan}
\textbf{Omni-o3} & \textbf{.564} \\
\bottomrule
\end{tabular}%
\end{adjustbox}%
\end{minipage}
}
\vspace{-6pt}

\subfloat[\textbf{Audio Visual Grounding}]
{%
\begin{minipage}[t]{\linewidth}
\centering
\begin{adjustbox}{width=\textwidth}
\begin{tabular}{
p{3.4cm}
p{.94cm}<{\centering} % AVE
p{.94cm}<{\centering}
p{.94cm}<{\centering} % FlickrSoundNet
}
\toprule
\multirow{2}{*}{Model} & \multicolumn{1}{c}{\textbf{AVE}} & \multicolumn{2}{c}{\textbf{SoundNet}} \\
\cmidrule(lr){2-2}
\cmidrule(lr){3-4}
& Acc & cIoU & AUC \\
\midrule

\rowcolor{gray!10}
\multicolumn{4}{l}{\textit{Task Specific MLLMs}}\\
Crab$_{\color{gray}{\it Mar\ 25}}$
&80.2&-   &-   \\
MeerKat$_{\color{gray}{\it Jul\ 24}}$
&N/A &88.4&67.9\\
\midrule

\rowcolor{gray!10}
\multicolumn{4}{l}{\textit{Omnimodal MLLMs}}\\
Qwen2.5-Omni$_{\color{gray}{\it Mar\ 25}}$
&75.7&06.6&09.3\\
Qwen3-Omni$_{\color{gray}{\it Sep\ 25}}$
&80.8&39.2&41.7\\
BuboGPT$_{\color{gray}{\it Jul\ 23}}$
&N/A &81.2&62.3\\
\midrule

\rowcolor{LightCyan}
\textbf{Omni-o3} &\textbf{85.1}&27.5&28.8 \\
\bottomrule
\end{tabular}%
\end{adjustbox}%
\end{minipage}
}
\end{minipage}
\end{minipage}
\vspace{-1em}
\label{tab:audio_visual}
\end{table*}

To validate our model, we conducted a comprehensive evaluation across three key dimensions: video modality, audio modality, and audio-visual modalities.

\noindent \textbf{\textit{Audio-Visual Capacity}}: 
% \\\\\\\\
To evaluate the comprehensive omnimodal capabilities of \model, we conduct experiments across 8  audio-visual benchmarks: AGAVQA~\cite{cao2025agav}, FlickrSoundNet~\cite{senocak2018learning}, AVE~\cite{zhou2025ave}, Video-MME~\cite{fu2025video}, WorldSense~\cite{hong2025worldsense}, DailyOmni~\cite{zhou2025daily}, Video-Holmes~\cite{cheng2025video}, and IntentBench~\cite{yang2025humanomniv2}. These datasets comprehensively assess the model's ability in multimodal reasoning and complex intent recognition.

\noindent \textbf{\textit{Video Capacity}}: 
% \\\\\\\\
To assess fine-grained spatial-temporal comprehension and temporal grounding, we conduct experiments on 9 benchmark datasets: VideoMMMU~\cite{hu2025video}, Vript-RR~\cite{yang2024vript}, Vript-ERO~\cite{yang2024vript}, VideoEspresso~\cite{han2025videoespresso}, LVBench~\cite{wang2025lvbench}, Charades-STA~\cite{sigurdsson2016hollywood}, V-STaR~\cite{cheng2025v}, TVG~\cite{wang2025timer1}, and FineVD~\cite{duan2025finevq}. These datasets are applied to evaluate the model in localizing and understanding visual concepts.

\noindent \textbf{\textit{Audio Capacity}}: 
% \\\\\\\\
To evaluate the model's auditory processing abilities, we conduct experiments on 6 benchmarks: QSpeech~\cite{wang2025qualispeech}, NISQA~\cite{mittag2021nisqa}, MMAU~\cite{sakshi2024mmau}, CommonVoice~\cite{ardila2019common}, VoiceBench~\cite{chen2024voicebench}, and MMSU~\cite{wang2025mmsu}. Specifically, QSpeech and NISQA are utilized to evaluate speech quality assessment, while MMSU is for complex audio reasoning.

\begin{table*}[!tbp]
\centering
\tabcolsep=2.8pt
\caption{\textbf{Comparison with leading MLLMs on 9 popular visual benchmarks.}
Bold numbers indicate the best performance in each category.
%
% Abbreviations: Arti. (Artifact), Temp. (Temporal).
}

% 左侧 Minipage 放 (a)
\begin{minipage}[t]{.604\linewidth}
\vspace{0pt} % <--- 关键修改：强制顶部对齐
\centering
\subfloat[\textbf{General \& Complex QA}]
{% <-- 注意这里的 % 极其重要，用于消除隐形空格，防止表格右偏
\begin{adjustbox}{width=\linewidth}
\begin{tabular}{
p{2.9cm}
p{1.3cm}<{\centering} % Video-MMMU
p{1.3cm}<{\centering} % Vript-RR
p{1.3cm}<{\centering} % Vript-ERO
p{1.3cm}<{\centering} % VideoEspresso-Test
p{1.3cm}<{\centering} % LVBench
}
\toprule
\multirow{2}{*}{Model} &\multicolumn{1}{c}{\textbf{VMMMU}} &\multicolumn{1}{c}{\textbf{VRR}} &\multicolumn{1}{c}{\textbf{VERO}} &\multicolumn{1}{c}{\textbf{Espresso}} &\multicolumn{1}{c}{\textbf{LVBench}} \\
\cmidrule(lr){2-2}
\cmidrule(lr){3-3}
\cmidrule(lr){4-4}
\cmidrule(lr){5-5}
\cmidrule(lr){6-6}
& All & Avg & @3 & All & All \\
\midrule

\rowcolor{gray!10}
\multicolumn{6}{l}{\textit{Closed-Source APIs}} \\
GPT-4o$_{\color{gray}{\it May\ 24}}$
&61.2&85.3&38.6&26.4&30.8\\
Gemini1.5Pro$_{\color{gray}{\it Dec\ 24}}$
&53.9&72.5&09.1&-   &33.1\\
\midrule

\rowcolor{gray!10}
\multicolumn{6}{l}
{\textit{Video MLLMs}} \\
Qwen2.5-VL$_{\color{gray}{\it Feb\ 25}}$
% &47.4&69.7&79.6&25.4&-   &45.3\\
&47.4&74.7&25.4&45.9&45.3\\
Qwen3-VL$_{\color{gray}{\it Oct\ 25}}$
&68.7&79.0&39.6&53.7&58.0\\
Open-o3-video$_{\color{gray}{\it Oct\ 25}}$ 
&52.3&74.3&32.8&48.9&44.5\\
VideoRFT$_{\color{gray}{\it May\ 25}}$ 
&51.4&77.7&30.6&48.2&40.7\\
VideoChat-R1$_{\color{gray}{\it Apr\ 25}}$ 
&51.1&72.7&35.1&49.4&40.8\\
\midrule

\rowcolor{gray!10}
\multicolumn{6}{l}
{\textit{Omnimodal MLLMs}} \\
Qwen2.5-Omni$_{\color{gray}{\it Mar\ 25}}$  
&53.2&72.4&32.1&46.4&41.5\\
Qwen3-Omni$_{\color{gray}{\it Sep\ 25}}$ 
&64.1&80.6&38.1&\textbf{51.4}&50.2\\
Ola$_{\color{gray}{\it Feb\ 25}}$  
&43.4&76.0&29.1&34.4&29.8\\
EchoInk-R1$_{\color{gray}{\it May\ 25}}$  
&35.2&73.1&23.9&49.1&38.7\\
Omni-R1$_{\color{gray}{\it May\ 25}}$  
&42.6&68.1&24.6&50.4&32.9\\
AV-Reasoner$_{\color{gray}{\it June\ 25}}$  
&41.6&72.7&28.4&35.8&29.4\\
HumanOmniV2$_{\color{gray}{\it June\ 25}}$  
&51.1&74.8&23.9&44.4&43.2\\
AVATAR$_{\color{gray}{\it CVPR\ 26}}$  
&53.7&75.7&36.6&52.2&38.4\\
\midrule

\rowcolor{LightCyan}
\textbf{\model}
&59.9&79.7&\textbf{40.3}&50.9&48.1\\
\bottomrule
\end{tabular}%
\end{adjustbox}%
}
\end{minipage}%
\hfill
% 右侧 Minipage 放 (b) 和 (c)
\begin{minipage}[t]{.38\linewidth}
\vspace{0pt} % <--- 关键修改：强制顶部对齐
\centering
\subfloat[\textbf{Quality Assessment}]
{%
\begin{adjustbox}{width=\linewidth}
\begin{tabular}{
p{3.2cm}
p{1.5cm}<{\centering}
p{1.5cm}<{\centering}
p{1.5cm}<{\centering} % FineVD 删减后的列宽适当调宽
}
\toprule
\multirow{2}{*}{Model} &\multicolumn{3}{c}{\textbf{FineVD}} \\
\cmidrule(lr){2-4}
& SRCC & Exist & Most \\
\midrule

\rowcolor{gray!10}
\multicolumn{4}{l}
{\textit{Task Specific MLLMs}} \\
FineVQ$_{\color{gray}{\it Dec\ 24}}$
&.883&91.9&65.1\\
\midrule

% \rowcolor{gray!10}
% \multicolumn{4}{l}
% {\textit{Video MLLMs}} \\
% Qwen2.5-VL$_{\color{gray}{\it Feb\ 25}}$
% &.438&27.6&34.8\\
% Qwen3-VL$_{\color{gray}{\it Oct\ 25}}$
% &-   &-   &-   \\
% \midrule

\rowcolor{gray!10}
\multicolumn{4}{l}
{\textit{Omnimodal MLLMs}} \\
Qwen2.5-Omni$_{\color{gray}{\it Mar\ 25}}$  
&.462&27.1&28.3\\
Qwen3-Omni$_{\color{gray}{\it Sep\ 25}}$ 
&.619&26.7&47.7\\
\midrule

\rowcolor{LightCyan}
\textbf{\model}
&\textbf{.743}&\textbf{31.0}&\textbf{43.6}\\
\bottomrule

\end{tabular}%
\end{adjustbox}%
}
\vspace{-7pt}

\subfloat[\textbf{Spatio-Temporal Grounding}]
{%
\begin{adjustbox}{width=\linewidth}
\begin{tabular}{
p{3.2cm}
p{1.01cm}<{\centering} % TVG
p{1.01cm}<{\centering} % Charades-STA
p{1.01cm}<{\centering}
p{1.01cm}<{\centering} % V-STaR
}
\toprule
\multirow{2}{*}{Model} &\multicolumn{1}{c}{\textbf{TVG}} &\multicolumn{1}{c}{\textbf{Cha.}} &\multicolumn{2}{c}{\textbf{V-STaR}} \\
\cmidrule(lr){2-2}
\cmidrule(lr){3-3}
\cmidrule(lr){4-5}
& mIoU & mIoU & AM & GM \\
\midrule

\rowcolor{gray!10}
\multicolumn{5}{l}
{\textit{Closed-Source APIs}} \\
GPT-4o$_{\color{gray}{\it May\ 24}}$
&-   &-   &26.8&38.2\\
% Gemini1.5Pro$_{\color{gray}{\it Dec\ 24}}$
% &-   &-   &-   &-   \\
\midrule

\rowcolor{gray!10}
\multicolumn{5}{l}
{\textit{Video MLLMs}} \\
Qwen2.5-VL$_{\color{gray}{\it Feb\ 25}}$
&14.8&43.6&19.3&22.4\\
Qwen3-VL$_{\color{gray}{\it Oct\ 25}}$
&28.7&56.0&18.8&21.1\\
Open-o3-video$_{\color{gray}{\it Oct\ 25}}$ 
&20.8&25.4&33.6&46.6\\
VideoRFT$_{\color{gray}{\it May\ 25}}$ 
&14.3&36.3&16.9&19.8\\
Time-R1$_{\color{gray}{\it Jun\ 25}}$ 
&29.2&58.1&27.8&36.0\\
\midrule

\rowcolor{gray!10}
\multicolumn{5}{l}
{\textit{Omnimodal MLLMs}} \\
Qwen2.5-Omni$_{\color{gray}{\it Mar\ 25}}$  
&15.0&48.9&17.3&20.5\\
Qwen3-Omni$_{\color{gray}{\it Sep\ 25}}$ 
&11.6&21.6&30.3&42.3\\
\midrule

\rowcolor{LightCyan}
\textbf{\model}
&\textbf{15.0}&23.3&27.1&35.0\\
\bottomrule

\end{tabular}%
\end{adjustbox}%
}
\end{minipage}

\label{tab:visual}
\end{table*}

\subsection{Results}
\myPara{Audio-Visual Benchmarks}
Table~\ref{tab:audio_visual} presents a comprehensive comparison of \model against 12 leading multimodal large language models (MLLMs) across five challenging audio-visual benchmarks. Overall, \model demonstrates overwhelming superiority, establishing a new state-of-the-art among open-source omnimodal models. We highlight its performance from two key perspectives:
(1) Robustness in standard contexts: On the AVE and Video-MME benchmarks, \model achieves overall scores of 85.1 and 75.7, respectively, outperforming the previous best open-source model, Qwen3-Omni (+4.3 on AVE, +5.2 on Video-MME).
(2) Effectiveness of deliberative reasoning in the complex scenario: On complex benchmarks requiring deep reasoning, such as WorldSense, Video-Holmes, and IntentBench, \model not only surpasses all open-source counterparts but also consistently eclipses powerful closed-source APIs like GPT-4o.
For instance, \model achieves 54.0 on Video-Holmes (vs. GPT-4o's 42.0) and 67.6 on IntentBench (vs. GPT-4o's 60.0).
Overall, the performance validates the efficacy of our nested deduction paradigm, which successfully bridges the gap between raw audio-visual perception and complex cognitive reasoning.

\myPara{Visual Benchmarks}
To evaluate whether our model maintains strong spatial-temporal grounding capabilities without suffering from modality interference, we assess \model on three pure video benchmarks: Vript-RR, TVG, and FineVD (Table~\ref{tab:visual}).
Overall, \model achieves highly competitive results, and its performance can be analyzed from two key perspectives:
(1) Superiority in video reasoning and grounding:
\model demonstrates robust visual comprehension, scoring 75.0 on Vript-RR (Whole-M) and 53.2 on the FineVD (Most) metric, substantially outperforming the strong omnimodal baseline Qwen3-Omni (72.4 and 47.7, respectively).
(2) Mitigating modality interference in dense grounding:
While specialized video-only models (\eg, Open o3-Video) still hold slight advantages in specific dense metrics like TVG mIoU (20.8 for Open o3-Video vs. 15.0 for \model), \model significantly narrows this gap compared to previous omnimodal architectures (\eg, Qwen3-Omni's 11.6).
In summary, these results demonstrate that our framework effectively preserves and even enhances fine-grained visual localization skills while jointly modeling audio and text.

\begin{table*}[!tbp]
\centering
\tabcolsep=2.8pt

\caption{\textbf{Comparison with leading MLLMs on 6 popular audio benchmarks.}
Bold numbers indicate the best performance in each category.
``Avg'' denotes the average PLCC on the respective benchmark.
%
% We report PLCC for quality assessment benchmarks.
}
% \vspace{-10pt}

\begin{minipage}[c]{.728\linewidth}
\subfloat[\textbf{General \& Complex QA}]
{%
\begin{adjustbox}{width=\textwidth}
\begin{tabular}{
p{3.2cm}
p{1.1cm}<{\centering} % MMAU
p{1.1cm}<{\centering} % MMSU
p{1.cm}<{\centering} % CommonVoice
p{.55cm}<{\centering}p{.55cm}<{\centering}p{.55cm}<{\centering}p{.55cm}<{\centering}p{.55cm}<{\centering}p{.55cm}<{\centering}p{.55cm}<{\centering}p{.55cm}<{\centering}p{.55cm}<{\centering}p{.55cm}<{\centering} % VoiceBench
}
\toprule
\multirow{2}{*}{Model} & \multicolumn{1}{c}{\textbf{MMAU}} & \multicolumn{1}{c}{\textbf{MMSU}} & \multicolumn{1}{c}{\textbf{CV15}} & \multicolumn{10}{c}{\textbf{VoiceBench}}\\
\cmidrule(lr){2-2}
\cmidrule(lr){3-3}
\cmidrule(lr){4-4}
\cmidrule(lr){5-14}
& All & All & Wer $\downarrow$&All&AE.&CE.&WV.&SD.&MM.&OB.&BB.&IF.&Adv. \\
\midrule

\rowcolor{gray!10}
\multicolumn{14}{l}{\textit{Closed-Source APIs}} \\
GPT-4o$_{\color{gray}{\it May\ 24}}$ 
&62.5&56.4&10.0&86.8&95.6&89.8&91.6&75.5&80.3&89.2&84.1&76.0&98.7\\
Gemini1.5 Pro$_{\color{gray}{\it Dec\ 24}}$  
&-   &60.7&-   &-   &-   &-   &-   &-   &-   &-   &-   &-   &-   \\
Gemini2.5 Pro$_{\color{gray}{\it May\ 25}}$  
&77.4&77.7&09.9&89.6&94.3&88.4&93.4&90.1&71.1&92.3&92.6&85.7&98.1\\
\midrule

\rowcolor{gray!10}
\multicolumn{14}{l}{\textit{Audio MLLMs}} \\
Qwen2-Audio$_{\color{gray}{\it Jul\ 24}}$  
&57.4&53.3&08.6&55.8&74.8&68.6&60.2&35.7&35.7&49.5&54.7&26.3&96.7\\
AudioReaonser$_{\color{gray}{\it Sep\ 25}}$  
&63.8&57.4&09.5&51.7&73.4&67.5&60.5&47.5&35.6&53.6&54.0&24.7&48.7\\
DeSTA2.5$_{\color{gray}{\it Jul\ 25}}$
&65.2&58.8&20.1&74.5&88.2&79.8&76.0&60.0&60.9&74.1&67.0&66.4&98.3\\
AFlamingo3$_{\color{gray}{\it NeurIPS\ 25}}$  
&72.4&61.4&07.4&38.1&42.6&35.8&37.1&36.0&41.0&57.4&52.1&29.2&11.7\\
\midrule

\rowcolor{gray!10}
\multicolumn{14}{l}{\textit{Omnimodal MLLMs}} \\
Qwen2.5-Omni$_{\color{gray}{\it Mar\ 25}}$  
&65.5&60.6&07.6&73.6&89.9&76.7&77.7&56.4&61.7&80.9&66.7&53.5&99.2\\
Qwen3-Omni$_{\color{gray}{\it Sep\ 25}}$  
&77.5&69.0&\textbf{06.1}&85.5&94.8&90.8&91.6&76.9&68.1&89.7&80.4&77.8&99.3\\
Ola$_{\color{gray}{\it Feb\ 25}}$ 
&67.1&62.6&13.1&59.4&82.4&59.4&63.8&33.8&46.0&67.9&51.1&39.6&90.8\\
\midrule

\rowcolor{LightCyan}
\textbf{Omni-o3}  
&75.6&\textbf{70.4}&07.9&57.7&85.0&49.2&66.6&46.4&31.7&33.6&63.5&47.0&96.0\\
\bottomrule

\end{tabular}%
\end{adjustbox}%
}
\end{minipage}
\hfill
\begin{minipage}[c]{.262\linewidth}
\centering
\subfloat[\textbf{Quality Assessment}]
{%
\begin{adjustbox}{width=\textwidth}
\begin{tabular}{
p{3.19cm}
p{.6cm}<{\centering} % QualiSpeech
p{.6cm}<{\centering} % NISQA
}
\toprule
\multirow{2}{*}{Model} &\multicolumn{1}{c}{\textbf{QS.}} &\multicolumn{1}{c}{\textbf{NIS.}}\\
\cmidrule(lr){2-2}
\cmidrule(lr){3-3}
& All & Avg \\
\midrule

\rowcolor{gray!10}
\multicolumn{3}{l}{\textit{Task Specific Models}} \\
ALLD$_{\color{gray}{\it Jan\ 25}}$ 
&-   &.907\\
Wav2Vec2$_{\color{gray}{\it Oct\ 20}}$ 
&-   &.910\\
QSpeech$_{\color{gray}{\it Jun\ 25}}$ 
&.660&-   \\
\midrule

\rowcolor{gray!10}
\multicolumn{3}{l}{\textit{Audio MLLMs}} \\
WavLLM$_{\color{gray}{\it Sep\ 24}}$ 
&.071&.006\\
SALMONN$_{\color{gray}{\it Apr\ 24}}$
&.084&.861\\
Qwen-Audio$_{\color{gray}{\it Nov\ 23}}$  
&.250&.771\\
Qwen2-Audio$_{\color{gray}{\it Jul\ 24}}$  
&.112&.768\\
\midrule

\rowcolor{gray!10}
\multicolumn{3}{l}{\textit{Omnimodal MLLMs}} \\
Qwen2.5-Omni$_{\color{gray}{\it Mar\ 25}}$  
&.104&.243\\
Qwen3-Omni$_{\color{gray}{\it Sep\ 25}}$  
&.319&.677\\
\midrule

\rowcolor{LightCyan}
\textbf{Omni-o3}  
&\textbf{.492}&\textbf{.815}\\
\bottomrule

\end{tabular}%
\end{adjustbox}%
}
\end{minipage}
\label{tab:audio}
\end{table*}

\myPara{Audio Benchmarks}
Table~\ref{tab:audio} details the evaluation on audio-centric benchmarks, focusing on speech quality assessment (QualiSpeech, NISQA) and multimodal speech understanding (MMSU).
Overall, \model achieves competitive results in audio processing, and its performance can be analyzed from two key perspectives.
(1) Fine-grained audio perception:
On the QualiSpeech benchmark, it attains an overall score of 0.492, outperforming both the audio models (\eg, Qwen2-Audio) and omnimodal models (\eg, Qwen3-Omni's 0.319).
Similarly, on NISQA, \model reaches an average score of 0.82.
(2) High-level audio reasoning:
Beyond low-level acoustic perception, \model excels in complex reasoning.
On the MMSU benchmark, it achieves an overall score of 70.4 (with 83.2 in Reasoning), outperforming GPT-4o (56.4) and getting close to the Gemini 2.5 Pro (77.7).
In summary, these results underscore that our model not only ``hears'' the audio but deeply comprehends its nuances, continuity, and naturalness.

\subsection{Visualization Results}
\begin{figure}[!t]
  \centering
    % \vspace{20pt}

  \includegraphics[width=\linewidth]{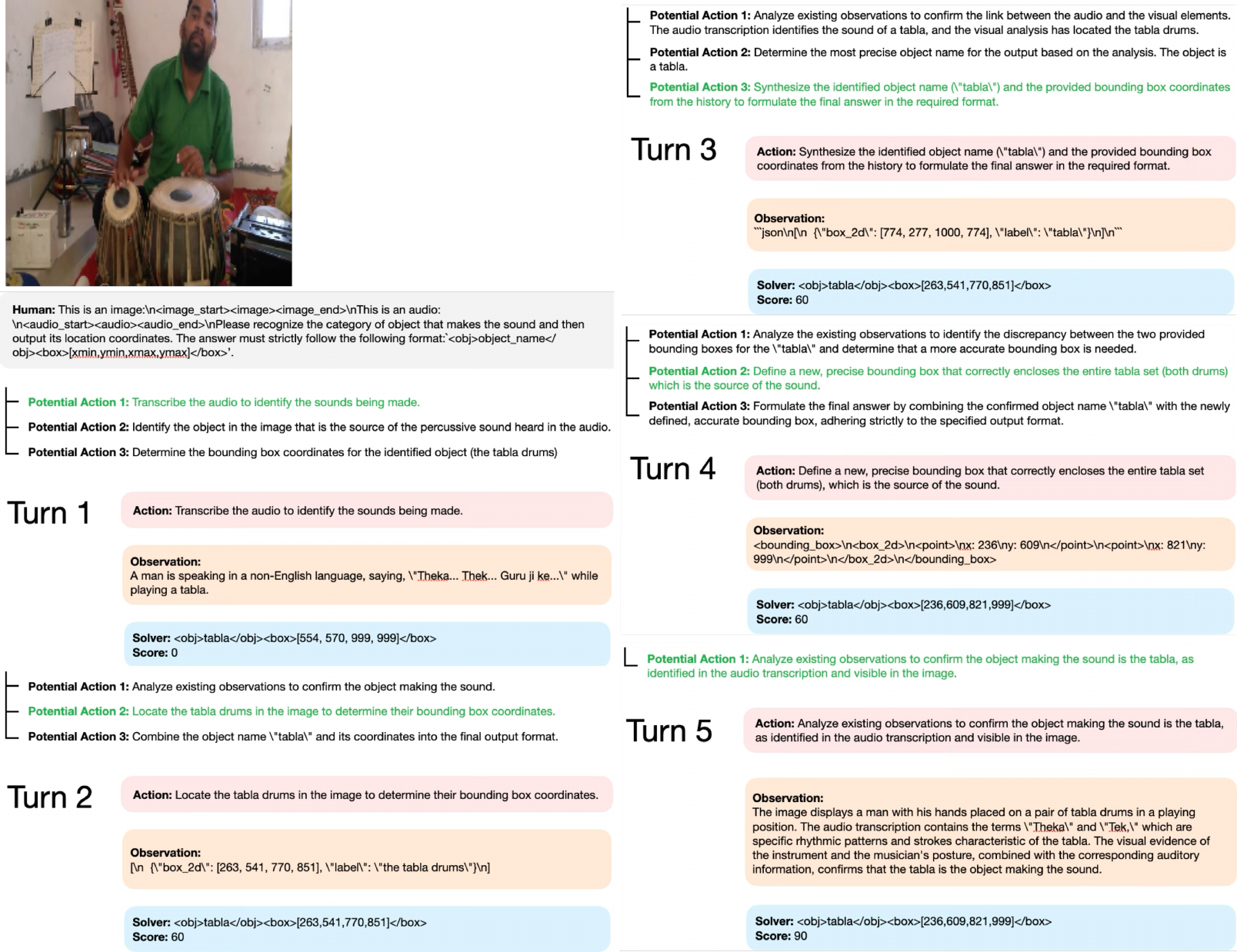}
  % \vspace{-20pt}
  \caption{\textbf{Qualitative visualization of \model's deliberative reasoning process.}
  We contrast two distinct cases:
  {(a) Single-Turn Reasoning} illustrates a straightforward audio-visual task where the model directly aligns an audio cue (barking) with visual objects (dogs) to output precise spatial coordinates in a single step.
  {(b) Multi-Turn Reasoning} showcases our model's advanced deduction capabilities on a complex video temporal grounding task.
  Instead of yielding a single prediction, \model\ dynamically constructs a reasoning tree.
  It iteratively proposes actions (\eg, identifying players, scanning sequences), evaluates intermediate observations, and effectively values its predictions until the target event is accurately localized.
  }
  % \vspace{-20pt}
  \label{fig:visualization}
\end{figure}

To intuitively demonstrate how \model bridges raw perception with system-2 cognitive capabilities, we present qualitative visualizations of its deliberative reasoning process in Figure~\ref{fig:visualization}.
We contrast two distinct reasoning paradigms to highlight the model's versatility and depth.

\myPara{Foundational Omnimodal Grounding}
Figure~\ref{fig:visualization}a illustrates a single-turn spatial reasoning task where the model is required to localize the source of an audio cue.
\model seamlessly aligns the auditory signal (barking) with the visual subjects (dogs).
It first generates a detailed, fine-grained observation of the scene (\ie, correctly identifying the breeds and positions of the two dogs), and subsequently outputs precise bounding box coordinates in a single deductive step.
This demonstrates the model's robust foundational omnimodal perception, which serves as the bedrock for more complex tasks.

\myPara{Dynamic Self-Correction in Long Videos}
Figure~\ref{fig:visualization}b showcases the model's advanced system-2 capabilities on a highly complex temporal grounding task within a dense hockey video.
Unlike traditional models that rely on a single-pass, feed-forward prediction, \model dynamically constructs a reasoning tree to navigate the long context. 
Initially ({Turn 1}), the model successfully identifies the target player (Chris Kunitz) but misjudges the exact temporal window of the shootout, resulting in a low internal valuation (Score: 0). 
Crucially, rather than failing silently, the model exhibits the Aha moment of {self-correction}. It initiates a sequence scan ({Turn 2}) and observes a shootout involving a different player, updating its partial understanding (Score: 60).
Recognizing the mismatch, it logically deduces the need to search for the subsequent attempt ({Turn 3}).
By proposing a new action to scan forward, it finally observes the correct event (Kunitz shooting, Niemi saving) and accurately localizes the temporal boundaries, achieving a high confidence valuation (Score: 90).

This multi-step trajectory perfectly mirrors the MCTS-driven search process it was trained on.
% %
It proves that \model has internalized the capacity for iterative exploration, valuation, and logical backtracking, enabling it to solve complex multimodal challenges that are insurmountable for standard linear reasoning models.

\section{Conclusion}
In this work, we introduced \model, a pioneering open-source omnimodal large language model designed to unify fine-grained audio-visual perception and complex system-2 cognitive reasoning.
Our framework integrates an intrinsic omni-skills bootstrapping stage with a novel two-stage nested deduction training paradigm, effectively mitigating modality interference and enabling interpretable, structured deliberative reasoning.
We further developed an MCTS-driven data engine to synthesize high-quality, rationale-grounded reasoning trajectories, which serves as a fundamental data infrastructure for training robust omnimodal models.
Experimental results across diverse audio, visual, and audio-visual benchmarks demonstrate that \model outperforms existing open-source counterparts, setting up a new milestone in omnimodal understanding and deep cognitive reasoning.

\noindent\textbf{Limitations.}
While \model exhibits strong generalization across diverse omnimodal tasks, several limitations remain.
Like most existing MLLMs, \model is still susceptible to hallucinatory content, particularly for highly noisy audio-visual contexts, which can lead to factually inconsistent narratives.
Besides, extending \model to realtime interaction remains underexplored.
We view \model as strong foundation for future works, enabling the exploration of more dynamic, real-time, and embodied omnimodal agents.


\bibliographystyle{splncs04}
\bibliography{main}
\end{document}